\documentclass[11pt,a4paper]{article}
\usepackage[hyperref]{acl2020}
\usepackage{times}
\usepackage{wrapfig}
\usepackage{latexsym}
\usepackage{url}

\usepackage{microtype}
\aclfinalcopy 

\usepackage{graphicx}
\usepackage{amsmath}
\usepackage{amsthm}
\usepackage{amssymb}
\usepackage{tikz}

\title{Topological Data Analysis for Word Sense Disambiguation}

\author{Michael Rawson \\
  \texttt{rawson@umd.edu} \\\And
  Samuel Dooley \\
  \texttt{sdooley1@umd.edu} \\\And
  Mithun Bharadwaj \\
  \texttt{mithun02@umd.edu} \\\And
  Rishabh Choudhary \\
  \texttt{rishchou@umd.edu} \\}

\date{}

\begin{document}
\maketitle
\begin{abstract}
    We develop and test a novel unsupervised algorithm for word sense induction and disambiguation which uses topological data analysis. 
    Typical approaches to the problem involve clustering, based on simple low level features of distance in word embeddings. 
    Our approach relies on advanced mathematical concepts in the field of topology which provides a richer conceptualization of clusters for the word sense induction tasks. 
    We use a persistent homology barcode algorithm on the SemCor dataset and demonstrate that our approach gives low relative error on word sense induction. 
    This shows the promise of topological algorithms for natural language processing and we advocate for future work in this promising area.
\end{abstract}

\section{Introduction}

Natural language processing applications have benefited from improved computational understanding of word senses. 
Various applications and downstream methods see gains in their performance when the system is able to disambiguate the specific sense of a word in its context, rather than using the surface-level word feature. 
A common way to accomplish this is to use manually compiled resources like WordNet.
WordNet is an example of a high-quality lexical resource that consolidates linguistic knowledge, but requires high levels of manual annotation and labor. 
While important, there are some drawbacks to the applicability to other languages, as well as the ability to update as language evolves. 

An alternative approach can use unsupervised approaches in the field of word sense disambiguation (WSD).
Here, the goal is to either identify the individual sense of a given word in the context of a larger phrase, or identify the number of senses for a given word that appears in a corpus. 
We focus on word sense induction (WSI), as many have previously, but propose a novel method to accomplish this.
Previous methods have been predicated on the existence of large corpra of background knowledge about senses, limiting the application of the techniques to English and explicitly requiring expensive, time-consuming processes to generalize to new languages and domains. 

Our approach relies on topological data analysis, a data analysis tool that uses advanced mathematical methods to look for high-level topological, or geometric features of a dataset. 
In unsupervised methods, like ours, there is little other information to draw on than where the words live in some embedding and how they relate to each other (how far apart they are).
Many of these unsupervised methods implicitly rely on topological or geometric properties of a connected graph of words to derive their senses.
However, these approaches do not use the most modern techniques to study these topological features, a problem which we aim to solve with our purely topological approach. 
We propose to use our algorithm of persistence diagrams to find salient topological features for individual words; we show that this approach can reliably recover word senses in an entirely unsupervised way. 

In this manuscript, we aim to answer the question: \emph{Can topological data analysis of word embeddings sufficiently recover word senses?}. 
In Section \ref{relwork} we describe the existing literature in both word sense disambiguation and topological data analysis. 
In Section \ref{approach}, we outline our proposed approach. 
In Section \ref{experiment}, we describe the setup of our experiments with analysis of the results towards our specific research question. 
Finally, in Section \ref{future}, we outline future work which is informed by the results from this project.

\section{Related Work}\label{relwork}

\subsection{Word Sense Disambiguation}
WSD was first framed as a computational task in the 1940s with Zipf's power-law theory.
Since then, the types of solutions to WSD can be binned into three categories: knowledge-based, supervised, and unsupervised. 
An example of the knowledge-based approach is \citet{mittal2015word} where semantic similarity is used to measure distances between words which theoretically ``measure" how much words are semantically similar. 
The approach taken in this example is different from our unsupervised approach, but the end goal is the same.
\citet{mittal2015word} support their claim of the appropriateness of (their specific) similarity measures by using real time implicit feedback from user.
An example of the supervised approach is \citet{carpuat2007improving} which uses a combination of different models: Naive Bayes, maximum entropy model, boosting, and kernel-PCA. 
This paper is the first of its kind to provide evidence that WSD can be used to improve the performance of statistical machine translation (SMT) tasks.
Like the previous example, the goal is the same as ours but the approach is different. 
\citet{carpuat2007improving} justify their claim using experiments comparing SMT against SMT + WSD across a variety of tasks and a variety of performance measures. 
The final approach is an unsupervised approach, like \citet{sinha2007unsupervised}.
These authors posit that graph-based centrality measures for word sense disambiguation can capture the necessary information. 
This approach is similar to ours in that there is an assumption that the geometry of a space carries some information about word senses. 
However, their work uses different similarity measures directly, while ours will use inferred measures through word embeddings.

\subsection{Topological Data Analysis}

This field has been studied for a long time, dating back to the 1940s (\citet{morse1940rank}, \citet{frosini1990distance}, and \citet{robins1999towards}).
These works focus on the theory of topology, but they were restricted in their applicability due to their lack of computation.
Then \citet{edelsbrunner2000topological} introduced a fast algorithm to compute a surface's topology, and this led to the focus of computational techniques in the field of topology. 
After the publication of this work, many subsequent work was focused on advancing the theory and computability of the shape of surfaces. 
For an abstract overview of these works see \citet{highlevel}.
For a history of persistent homology in TDA see \citet{perea2018brief}.
For more mathematical and technical descriptions of the field see \citet{ghrist2018homological}, \citet{edelsbrunner2012persistent}, and \citet{chazal2016high}.

There are only six published works at the intersection of NLP and TDA (\citet{wagner2012computational}, \citet{zhu2013persistent}, \citet{michel2017does}, \citet{sami2017simplified}, \citet{temvcinas2018local}, \citet{savle2019topological}). 
The first concrete example of a successful application of TDA to NLP comes from \citet{wagner2012computational} which demonstrates the difficulties and possibilities for computational topology to analyze the similarities within a collection of text documents.
\citet{temvcinas2018local} argues for applicability of persistent homology to lexical analysis using word embeddings. 
This paper aims towards the same goal as ours, but does so using a slightly different TDA method, which we believe we can improve upon.
\citet{savle2019topological} applies topological data analysis (TDA) to entailment, with an improvement of accuracy over the baseline without persistence. 

\section{Proposed Approach}\label{approach}

Most word sense disambiguation methods approach the problem by examining how words are related to each other in some embedded space. 
These approaches develop a concept word sense by doing clustering or looking at other elements of the shape of the data. 
In our approach, we follow this core idea, but we expand on it through topological data analysis (TDA).
We describe at a high-level, what TDA is and how our algorithm works. 

In general data analysis, we often want to understand the shape or topology of any dataset we are given, where we do not know this information a priori.
Often, we do not know what the shape of the data is, but believe the data has been sampled from some underlying surface or manifold lesser in dimension than the ambient space.
The shape of this surface is interesting, but difficult to analyze so we focus on topological features which are much simpler. 
For example we check if the surface has holes or even multiple connected components which will correspond to clusters of the data.

\begin{figure}
\centering
\includegraphics[width=\linewidth]{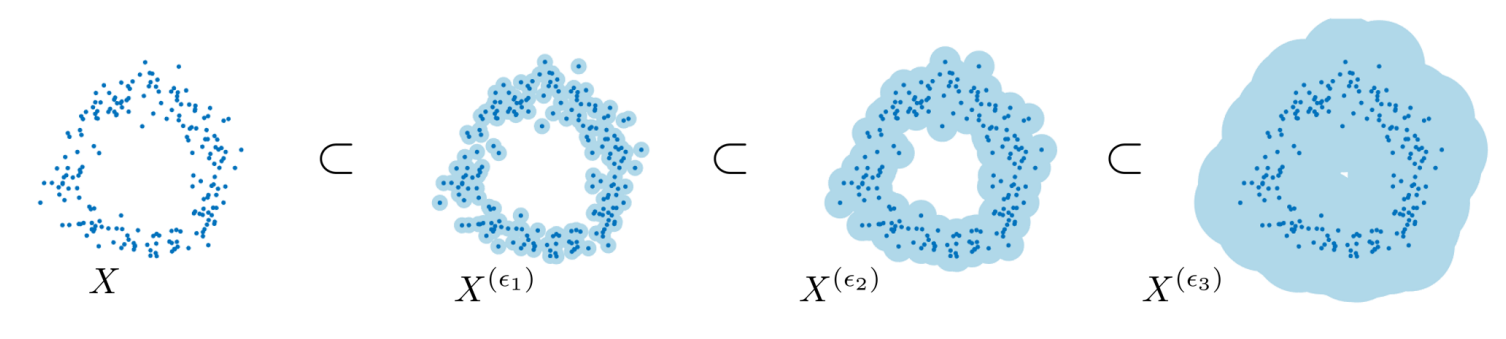}
    \caption{Example dataset $X$ (far left) which appear to be noisily sampled from a circle. In topological data analysis, we hope to be able to say that these data are from some object that are all connected but have a hole in the middle. In the barcode algorithm, we can do this by increasing the size of $\epsilon$-balls around the points and looking at the birth and death of topological features such a connected components or holes. Figure attributed to \citet{perea2018brief}.}
\label{fig1}
\end{figure}

To formalize this question, suppose you are given some (finite) data $X$ which you believe come from some surface\footnote{technically some embedded manifold} $\mathbb{X}$. 
The critical question is whether we can infer the shape\footnote{technically the homology} (number of holes, compontents, etc.) of $\mathbb{X}$ from the given data $X$.
Consider the data point cloud on the left side of in Figure \ref{fig1}. 
Intuitively we see that this data comes from noisily sampling a circle because there is some hole in the middle of the data.
But how would one algorithmically detect this for any given data?
The most common way to do this is to use a concept called persistence. 
The core idea of persistence is to see how the union of balls around each point becomes connected as the balls grow in radius.

\begin{figure}
    \centering
    \includegraphics[width=7.5cm, height=4cm]{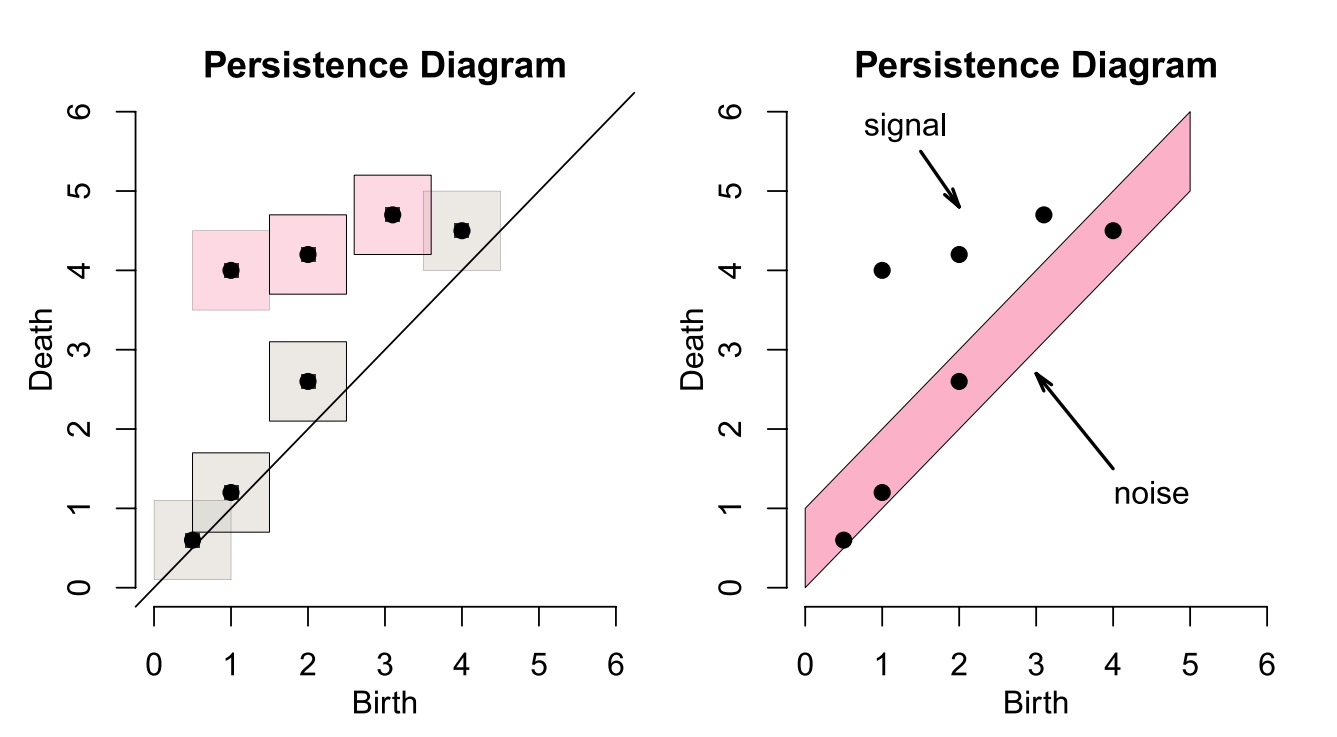}
    \caption{As we increase $\epsilon$, i.e., the size of the ball around each point in $X$ grows, we observe each connected component, noting when that connected component began (was born) and when it ended (died). We plot these on the persistence diagrams. Each dot represents on connected component which appeared as we were increasing $\epsilon$. We hypothesize that those components which did not live for a long time are just noise. These correspond to the components plotted close to the $y=x$ line. Therefore, we consider the components in the red zone on the right to be noise and those above that are signal about the topology of the data.}
    \label{fig:my_label}
\end{figure}

In Figure \ref{fig1}, we see that as we grow an $\epsilon$-ball around each point, the union of balls becomes more connected until we have all balls connected and 1 hole in the middle of the union. In persistent homology, we examine how these connections evolve overtime (time being the ball radius increasing). 
For the different discriminators of different shapes (connected components, holes, etc), we can observe their birth and death as we increase $\epsilon$. 
In Fiugre \ref{fig:my_label}, we see that we can plot the stages of birth and death for different topological features. 
Those features which die around the time they are born (are plotted along the line $y=x$), we can consider these features to be noise. 
By focusing on those discriminators which have a long life, i.e., persist, we can begin to understand the shape of our data; this is done by taking a noise tolerance around the $y=x$ line and considering those that live outside that threshold (see Figure \ref{fig:my_label} (right)).

Once we have the word embeddings, we will run a topological data analysis algorithm to test our Research Question. 

Recall, that the study of topology examines the surfaces that are generating the data. For instance, we assume that our word embeddings are produced from some underlying manifold that has particular structures of interest. For instance, word embeddings make ``similar'' words cluster together in the embedding space. We note that the definition of similar is somewhat nebulous. In topology, we call these clusters \emph{connected components} because if we were to connect each of the points/word embeddings to any other point that is say $\epsilon$ away, then they would form one connected piece. This point can be seen in Figure \ref{fig1}. Recall that persistent homology is the study of the evolution of these topological features as we increase $\epsilon$. 

\subsection{Barcode Algorithm}
\label{barcode-section}
We explain the barcode algorithm for the 0th persistent homology\footnote{See: https://www.math.upenn.edu/~ghrist/preprints/barcodes.pdf. Our description is inherently reductive because of the constraints on the length of the assignment.}. We start with a list of $N$ data points, $X$, each in $d$ dimensional Euclidean space. Then we calculate the distance between every pair of points, which corresponds to the $e$ at which an edge is added between the pair in $VR_e$. Take the list of distances, $E$, and sort it in increasing order. Remove duplicate elements in $E$ and call it $D$. We build a matrix, $M$, with a column for each edge and a row for each vertex in $VR_\infty$ (the complete graph). Set $M(i,j)$ to $t^a$ when $i$ is a vertex of edge $j$ and where $a$ is the index of edge $j$ in $D$ which ranges from $1$ to the length of $D$. All other entries of $M$ are set to 0. We make the entries to be polynomials of variable $t$ with coefficients restricted to 0 or 1. The next step is to column reduce this matrix so that what remains is lower triangular. Then the remaining nonzero diagonal entries, say $t^b$, correspond to the barcode or interval $(0,b)$. With this list of intervals, the algorithm is complete.

The output for this algorithm will be a diagram like in Figure \ref{fig:my_label}. This diagram will tell us which are topological features of the word embedding space that are meaningful. In Figure \ref{fig:my_label} there are the three signals which fall above the noise-thresholded diagonal. With the topological information output from the barcode algorithm, we can look at words that have multiple senses. The essential quesiton we ask will be, if we remove this word/point from the embedding, does the topology change. To think of this more concretely, consider the word ``bank'' which has a financial and a alluvial meaning, each with their own cluster of similar words. However, ``bank'' falls between them and so if it is removed, the connection between the two clusters will be broken and one connected component will become two. We shall compute these changes for various words to test our Research Question. 

\section{Experiments and Results}\label{experiment}

To answer our Research Questions, we used the SemCor dataset to train word2vec on various embedding dimension sizes. 
Then, we used the barcodes described in Section \ref{approach} to understand the topology, or shape, of the word embedding space.
We  analyze the results of this algorithm by finding the topological components that each word belongs to, as output from the barcodes algorithm. 
The performance of the algorithm varies with the dimension of the embeddings and the number of neighboring words being considered. 
We will perform the computations on the CSCAMM servers for mass parallelization.

\textbf{Data}: Our baseline results have been computed on the Semcor dataset with the word2vec embedding training routine.

\textbf{Evaluation pipeline}:
\begin{center}
    \resizebox{\linewidth}{.2\linewidth}{
\begin{tikzpicture}[x=0.75pt,y=0.75pt,yscale=-1,xscale=1]

\draw   (143.5,57) -- (508.5,57) -- (508.5,225) -- (143.5,225) -- cycle ;
\draw    (92.5,139) -- (140.5,139.96) ;
\draw [shift={(142.5,140)}, rotate = 181.15] [color={rgb, 255:red, 0; green, 0; blue, 0 }  ][line width=0.75]    (10.93,-3.29) .. controls (6.95,-1.4) and (3.31,-0.3) .. (0,0) .. controls (3.31,0.3) and (6.95,1.4) .. (10.93,3.29)   ;

\draw    (507.5,141) -- (547.5,141) ;
\draw [shift={(549.5,141)}, rotate = 180] [color={rgb, 255:red, 0; green, 0; blue, 0 }  ][line width=0.75]    (10.93,-3.29) .. controls (6.95,-1.4) and (3.31,-0.3) .. (0,0) .. controls (3.31,0.3) and (6.95,1.4) .. (10.93,3.29)   ;

\draw   (155,91) -- (368.5,91) -- (368.5,209) -- (155,209) -- cycle ;
\draw    (370.5,141) -- (391.5,141) ;
\draw [shift={(393.5,141)}, rotate = 180] [color={rgb, 255:red, 0; green, 0; blue, 0 }  ][line width=0.75]    (10.93,-3.29) .. controls (6.95,-1.4) and (3.31,-0.3) .. (0,0) .. controls (3.31,0.3) and (6.95,1.4) .. (10.93,3.29)   ;

\draw    (3,108.5) -- (91,108.5) -- (91,175.5) -- (3,175.5) -- cycle  ;
\draw (47,142) node  [align=left] {Train Word \\Embedding \\on Corpus};
\draw    (549.5,120) -- (618.5,120) -- (618.5,166) -- (549.5,166) -- cycle  ;
\draw (584,143) node  [align=left] {Compute\\Errors};
\draw    (161,118) -- (237,118) -- (237,164) -- (161,164) -- cycle  ;
\draw (199,141) node  [align=left] {Get Local \\Homology};
\draw    (277,97) -- (359,97) -- (359,185) -- (277,185) -- cycle  ;
\draw (318,141) node  [align=left] {Count The \\Number of\\Significant\\Barcodes};
\draw    (392,107.5) -- (488,107.5) -- (488,174.5) -- (392,174.5) -- cycle  ;
\draw (440,141) node  [align=left] {Compare to\\Ground Truth\\Sense Count};
\draw (201,74) node  [align=left] {For each Word};
\draw (203,193) node  [align=left] {{\footnotesize Our Algorithm}};
\draw    (237,141) -- (275,141) ;
\draw [shift={(277,141)}, rotate = 180] [color={rgb, 255:red, 0; green, 0; blue, 0 }  ][line width=0.75]    (10.93,-3.29) .. controls (6.95,-1.4) and (3.31,-0.3) .. (0,0) .. controls (3.31,0.3) and (6.95,1.4) .. (10.93,3.29)   ;

\end{tikzpicture}
}
\end{center}

The persistent homology barcode algorithm discussed in Section \ref{approach} returns the lifespan of the connected components (barcodes) which we look at those which are statistically significant (two standard deviations away from the mean). As such, we define the number of senses to be those topological components that are above this noise zone, as depicted in Figure \ref{fig:my_label}. The result is considered our predicted number of senses for the target word. We compute absolute and relative errors when compared to the ground truth number of senses.

\textbf{Metric}: We will compare the number of senses calculated for each word with the ground truth from the annotated dataset. We will compute the number of senses based on the topological notions described at the end of Section \ref{barcode-section}. We shall look at the topological features of a word with multiple sense. We will then compute how the topology changes when that word is omitted. The average relative error, comparing word by word, will be the measure of success with a perfect match of average relative error = 0. We will measure the success against the hyperparameters $\delta$, the locality radius, and $\epsilon$, the noise sensitivity. That is if $g$ is the ground truth vector of number of word senses per word with length $n$, and $\Tilde{g}$ is our approximate, then the average relative error is 
$$e_{\delta,\epsilon}
=\frac{1}{n}\sum_{i=1}^n \frac{|(g)_i-(\Tilde{g}_{\delta,\epsilon})_i|}{(g)_i}.$$

This measure is appropriate for our Research Questions. The above measures the relative difference in word sense that our barcode algorithm predicts versus what are given as ground truth. This will measure effectively how this particular topological approach worked with recovering word senses. However, we note that even if this measure is large, there might be useful information encoded with the barcode algorithm. Therefore, we will also perform qualitative analyses to derive reasons for the results.

\textbf{Ground truth}: The datasets we have is annotated with the WordNet senses and counts which appear in them. We compare our predicted sense count with the ground truth by using both relative error and absolute error.

\textbf{Results\footnote{Repository for our implementation: \url{https://github.com/michrawson/TDA_word_embeddings}}}: We report a subset of our results in Table \ref{tab:my_label}. 
Our algorithm has two main variables that we can tune. 
First, we can tune the dimension of our embedding space; in this case, it is how many dimensions we train word2vec to have. 
The other main variable is the number of neighbors we consider when we are computing the topology of a word.
Over an exhaustive and representative sample of these hyperparameter combinations, we find that we are able to recover the number of senses very well. 
Our lowest relative error that we were able to achieve when we consider all words (with between 2 and 19 senses) is 0.518.
The hyperparameters which achieve this optimal performance are an embedding dimension of 500 and a neighbor count of 200. 

\begin{table}[]
    \centering
    \resizebox{\linewidth}{!}{%
    \begin{tabular}{|c|c||c|c|}
    \hline
    \multicolumn{4}{|c|}{Words with 2-9 Senses} \\
    \hline
    \begin{tabular}[c]{@{}c@{}}Embedding \\ Dimension\end{tabular} & \begin{tabular}[c]{@{}c@{}}Num Neighbors\\ Considered\end{tabular} & \begin{tabular}[c]{@{}c@{}}Relative \\ Error\end{tabular} & \begin{tabular}[c]{@{}c@{}}Absolute\\ Error\end{tabular} \\ \hline\hline
    500                                                            & 100                                                                & 0.9023                                                    & 2.723                                                    \\ \hline
    500                                                            & 50                                                                 & 0.4178                                                    & 1.660                                                    \\ \hline
    500                                                            & 25                                                                 & 0.4114                                                    & 2.202                                                    \\ \hline
    100                                                            & 100                                                                & 0.8030                                                    & 2.447                                                    \\ \hline
    100                                                            & 50                                                                 & 0.4331                                                    & 1.745                                                    \\ \hline
    100                                                            & 25                                                                 & 0.4348                                                    & 2.287                                                    \\ \hline
    \end{tabular}
    }
    \vspace{1em}
    \resizebox{\linewidth}{!}{%
\begin{tabular}{|c|c||c|c|}
    \hline
    \multicolumn{4}{|c|}{Words with 10-19 Senses} \\
    \hline
\begin{tabular}[c]{@{}c@{}}Embedding \\ Dimension\end{tabular} & \begin{tabular}[c]{@{}c@{}}Num Neighbors\\ Considered\end{tabular} & \begin{tabular}[c]{@{}c@{}}Relative \\ Error\end{tabular} & \begin{tabular}[c]{@{}c@{}}Absolute\\ Error\end{tabular} \\ \hline\hline
1000                                                           & 200                                                                & 0.2129                                                    & 2.907                                                    \\ \hline
1000                                                            & 100                                                                 & 0.4609                                                    & 6.511                                                    \\ \hline
500                                                            & 200                                                                 & 0.1897                                                    & 2.6511                                                    \\ \hline
500                                                            & 100                                                                & 0.4712                                                    & 6.6279                                                    \\ \hline
100                                                            & 200                                                                 & 0.2068                                                    & 3.0465                                                    \\ \hline
100                                                            & 100                                                                 & 0.4855                                                    & 6.5116                                                    \\ \hline
\end{tabular}
    }
    \caption{We report the relative and absolute error of the persistent homology barcode algorithm on the number of word senses for two sets of words: (1) words with 2-9 senses, and (2) words with 10-19 senses.}
    \label{tab:my_label}
\end{table}

However, we will also separate these results into two components: those words with few senses and those with many.
The relative error for words with few senses will be higher than the relative error for words with more senses. 
This is because the denominator of the relative error is lower so any errors (which have to be integer errors) will count for more. 
Therefore, in Table \ref{tab:my_label}, we break these two groups apart and report results for both separately. 

\begin{figure}
    \centering
    \includegraphics[width=\linewidth]{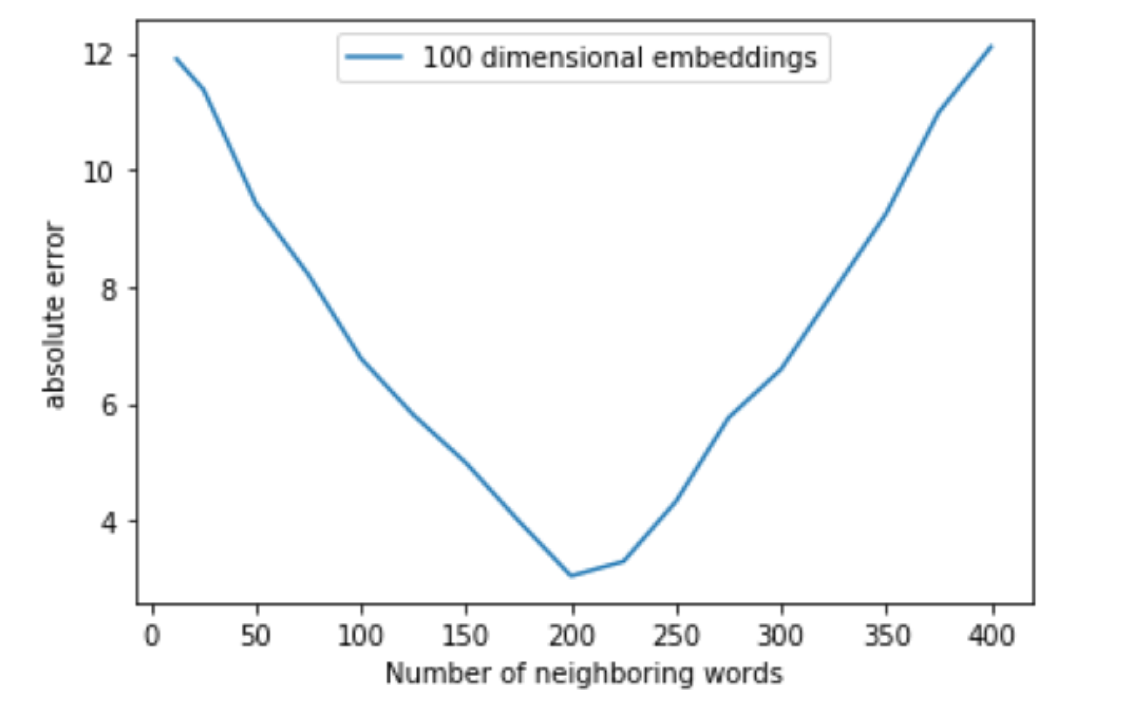}
    \caption{We plot the absolute for words with 10-19 senses. We fix the word2vec embedding at 100 and vary the number of neighbors we consider. We observe a clear minimum around 200 neighbors. We further state that this trend is persistent across multiple embedding dimensions.}
    \label{fig:my_label2}
\end{figure}

We see that for words with fewer senses, the optimal achievable relative error is 0.411 with word2vec embedding dimension of 500 and the number of neighbors considered is 25. 
For words with more senses, the optimal achievable relative error is 0.1897 with word2vec embedding dimension of 500 and number of neighbors considered is 200. 
We see that this is an interesting difference between the two groups. 
The relative error for words with fewer senses is positively correlated with the number of neighbors considered, i.e., as the number of neighbors goes down, so does the relative error. 
However, with words that have many senses, we see that there is a more complex interplay at work. We would expect that if we consider very few neighbors, we will not have enough information required to cluster all the senses and if we consider a very high number of neighbors, we expect to include a lot of irrelevant information for accurate clustering. Our intuition is that in both the above mentioned cases the error will be high.
In Figure \ref{fig:my_label2}, we see that 200 neighbors is the clear minimum of words to be included to minimize the error of this system. 

\section{Future Work}\label{future}
Further we would like to implement the algorithm on different datasets such as SEMEVAL-2013, and SemCor+OMSTI (One Million Sense Tagged Instances) using different word embeddings like GloVe and FastText to observe the impact on the results.

We would also like to test our algorithm with pseudowords. For a word that has multiple meanings or senses, we can break that word up into its component senses, create new dummy words, and replace the original word with the respective dummy word. For example, if the word ``foo'' has two meanings, we can create words ``foo\$1'' and ``foo\$2'' and replace ``foo'' with the respective pseudoword we just created.
Once we have done this, perhaps on some subset of the words in the datasets, we will replace the words in the corpora, retrain the embeddings, and rerun the barcode algorithm and infer the appropriateness of topological data analysis in word sense disambiguation/induction.

\bibliography{refs}
\bibliographystyle{acl_natbib}

\end{document}